\PassOptionsToPackage{table}{xcolor}
\documentclass{article} % For LaTeX2e
\usepackage{iclr2027_conference,times}

\usepackage{amsmath,amsfonts,bm}

\def\eqref#1{equation~\ref{#1}}
\def\1{\bm{1}}

\DeclareMathAlphabet{\mathsfit}{\encodingdefault}{\sfdefault}{m}{sl}
\SetMathAlphabet{\mathsfit}{bold}{\encodingdefault}{\sfdefault}{bx}{n}

\usepackage[table]{xcolor}
\definecolor{mycolor}{RGB}{215, 215, 215}
\definecolor{myrowcolor}{RGB}{230, 230, 230}
\definecolor{mycolor2}{RGB}{225, 225, 255}

\usepackage{hyperref}
\usepackage{url}
\usepackage{graphicx}
\usepackage{caption}
\usepackage{booktabs}
\usepackage{multirow}
\usepackage{wrapfig}
\usepackage{placeins}
\usepackage{amsmath}
\usepackage{amssymb}

\title{Motion-Guided Contrastive Regularization for Latent World Models}

\author{
Yiwen Chen$^{1}$\thanks{Work done during an internship at Qualcomm.} \quad
Qiqi Hou$^{2}$ \thanks{Corresponding author.} \quad
Jisoo Jeong$^{2}$ \quad
Yinhao Zhu$^{2}$ \quad
Huaizu Jiang$^{1}$ \quad
Amir Said$^{2}$ \\[2pt]
\mdseries
\begin{tabular}{@{}l@{\hspace{1.5em}}l@{}}
$^{1}$Northeastern University & {\small\texttt{\{chen.yiwe,h.jiang\}@northeastern.edu}} \\
$^{2}$Qualcomm AI Research \thanks{Qualcomm AI Research is an initiative of Qualcomm Technologies, Inc.} & {\small\texttt{\{qhou,asaid,jisojeon,yinhaoz\}@qti.qualcomm.com}}
\end{tabular}
}

\newif\ifshowtodos
\showtodosfalse

\newcommand{\old}[1]{}

\iclrfinalcopy % Uncomment for camera-ready version, but NOT for submission.

\begin{document}

\maketitle

\begin{abstract}
JEPA-style world models learn action-conditioned latent dynamics without reconstructing observations, but distributional regularization alone does not ensure that their representations preserve information needed for control. We identify \emph{static collapse}, a failure mode in which the encoder preserves episode-specific static content while suppressing manipulation-relevant state changes. Because static content is temporally predictable yet diverse across episodes, this failure can persist despite regularizers such as SIGReg. A controlled study on OGBench-Cube shows that introducing randomized backgrounds substantially degrades planning performance while leaving the underlying manipulation task unchanged. To address this problem, we propose \emph{motion-guided contrastive regularization}. Our approach compares predicted latent transitions with transitions to candidate observations from the same episode, reducing the usefulness of static background identity for distinguishing action outcomes. Optical-flow similarities provide soft contrastive targets, allowing candidates with similar motion to receive similar supervision. Combined with latent prediction and SIGReg, our objective encourages representations to preserve within-episode state distinctions and similarities between related transitions. Our method improves planning success over LeWorldModel (LeWM) from 52\% to 78\% on OGBench-Cube with randomized backgrounds and from 14\% to 54\% on LIBERO-Object, while maintaining comparable performance on four other visual-control benchmarks.
\end{abstract}

% JEPA-style world models learn action-conditioned latent dynamics without reconstructing observations, but their representations can discard information needed for control. We demonstrate \emph{static collapse}: the encoder preserves episode-specific static content while suppressing changes relevant to manipulation. Such content is temporally predictable yet diverse across episodes, allowing this failure to persist despite the diversity encouraged by regularizers such as SIGReg. A controlled study on OGBench-Cube shows that introducing randomized backgrounds sharply degrades planning performance while leaving the underlying manipulation task unchanged. To address this problem, we introduce a \emph{motion-guided contrastive loss}. We sample candidate observations from the same episode and compare their transitions from a shared reference frame. Motion-weighted optical-flow distances define a soft target distribution, which supervises the distribution of cosine similarities between the corresponding latent differences. Added to the existing prediction and regularization objectives, this loss encourages latent transitions to preserve the similarity relationships of observed motion. Our method improves planning success over LeWM from 52\% to 78\% on OGBench-Cube with randomized backgrounds and from 14\% to 54\% on LIBERO-Object, while maintaining comparable performance on four other visual-control benchmarks.
\begin{figure}[t]
 \centering
\includegraphics[width=\linewidth]{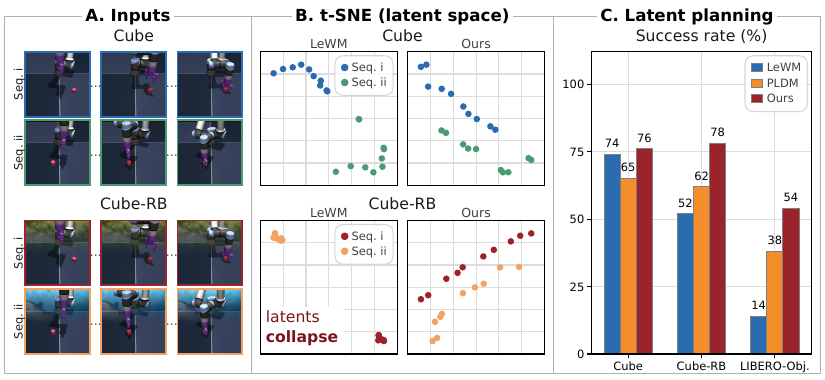}
 \caption{\textbf{Static collapse and our solution.}
\textbf{(A)} We train JEPA world models end-to-end on \emph{OGBench-Cube} (Cube) and a variant with randomized backgrounds (Cube-RB). In Cube-RB, one background image is sampled per episode and held fixed throughout it. Only distant pixels (depth $\ge 1.0$\,m) are replaced, so the robot, cube, and dynamics are unchanged and only task-irrelevant background variation is added.
\textbf{(B)} t-SNE visualizations of the encoder embeddings, where each point is one input frame and each color one episode. On Cube-RB, LeWorldModel \citep[LeWM;][]{lewm} maps distinct states within each episode to tightly clustered latents, whereas our motion-guided contrastive regularization preserves within-episode state variation.
\textbf{(C)} Planning success of LeWM, Planning with Latent Dynamics Models \citep[PLDM;][]{pdlm}, and our method. LeWM and our method perform similarly on Cube, but LeWM's success drops on Cube-RB and is low on LIBERO-Object. Our method achieves the highest success of the three in all settings.}
\label{fig:teaser}
\end{figure}

\section{Introduction}
\label{sec:introduction}

% \qiqi{test} \yinhao{test} \amir{test} \jisoo{test} \hj{test} \TODO{test}

% 1. What's world model.
% 2. what's jepa
% 3. Jepa is easy to collapse

% In action-conditioned world models like LeWorldModel (LeWM) ~\cite{}, this predictive formulation allows the model to discard unnecessary visual noise.

World models have attracted considerable attention due to their wide applications in robotics and autonomous driving. By learning environment dynamics to predict the consequences of actions, they enable agents to plan and learn behaviors through imagined experience ~\citep{JEPA-intro,WMTrainingAI,transformerWM}. However, learning representations that capture critical state information without being overwhelmed by high-dimensional visual details is challenging. 

Joint Embedding Predictive Architectures (JEPAs) offer a promising solution by predicting target representations from context, completely bypassing pixel-level reconstruction \citep{ijepa}. It allows an encoder to map observations into latent states while a predictor estimates future embeddings based on observation history and actions \citep{lewm}.  However, abandoning pixel reconstruction introduces a critical vulnerability to representational collapse, where the unconstrained objective trivially maps all observations to a constant embedding ~\citep{lewm}.

Considerable effort has been devoted to learning non-collapsing representations in predictive world models. Early approaches to prevent representation collapse primarily focused on constraining encoder updates, either through moving-average target encoders~\citep{ijepa, vjepa} or by freezing large pretrained visual encoders~\citep{dinowm, vjepa2}. These methods stabilize prediction targets and achieve strong empirical results. However, they do not explicitly force representations to distinguish between different action outcomes, and frozen-encoder approaches additionally incur significant computational overhead from their large pretrained backbones. Alternatively, other approaches prevent collapse by supervising action information \citep{sensormoterworld,delta-jepa, pdlm}. By employing inverse dynamics to recover actions from latent transitions or differences, these methods successfully establish meaningful state representations. Nevertheless, their supervisory signal is inherently focused on recovering the executed action itself, rather than explicitly distinguishing a predicted future outcome from competing alternatives. More recent efforts have turned to regularizing the embedding distribution itself \citep{vicreg,lejepa,lewm}. 
For instance, by employing Sketched Isotropic Gaussian Regularization (SIGReg) to encourage isotropic Gaussian latent spaces, LeWorldModel~\citep[LeWM;][]{lewm} trains a lightweight ViT that achieves results comparable to DINO-WM~\citep{dinowm} while planning more than $50\times$ faster (Table~\ref{tab:main_results}).

However, SIGReg constrains embedding statistics without specifying which information to preserve, allowing static backgrounds to dominate while control-relevant state changes are suppressed. We term this failure \textbf{static collapse}. Figure~\ref{fig:teaser} illustrates it on OGBench-Cube-RB, constructed by replacing distant background pixels in OGBench-Cube~\citep{ogbench} while preserving the robot and cube. LeWM's t-SNE embeddings cluster tightly within each episode despite changes in manipulation state, and its planning success drops substantially relative to the original dataset. These results show that statistical regularization alone does not ensure preservation of the state distinctions needed for control.

% 6. introduce our method
In this paper, we introduce motion-guided contrastive regularization for JEPA-style world models. Our approach is motivated by two key observations. First, distributional regularization, e.g., SIGReg, alone does not ensure that representations preserve state differences within the same environment. We therefore introduce a contrastive objective that compares the predicted action outcome with the observed next state and alternative states from the same episode. Because these observations share the same static scene, background identity alone cannot resolve the comparison, encouraging the model to capture changes in the robot and manipulated objects. Second, alternative states are not equally different from the observed outcome. Treating them all as negatives can push apart states that reflect similar manipulation outcomes, while measuring their similarity through pixel differences can be sensitive to irrelevant appearance variations. We therefore use optical flow to guide the contrastive objective, assigning greater similarity to alternatives whose motion resembles the observed transition. Together, these two ideas encourage the model to distinguish meaningful state changes while preserving similarities between related outcomes. Combined with latent prediction and SIGReg, our approach targets static collapse without relying on pixel reconstruction losses or auxiliary action-decoding objectives.

% In this paper, we introduce motion-guided contrastive regularization for JEPA-style world models. Our approach is motivated by two key observations. First, rather than relying solely on SIGReg to prevent representation collapse, we introduce a contrastive objective that leverages temporal dynamics. Given the current observation and action, the true next observation is treated as a positive sample, while other observations from the same episode serve as negative samples. The model is trained to assign a smaller latent-space distance to the positive pair than to the negative pairs. Because all samples share the same static scene, background appearance alone is insufficient to identify the correct future observation, forcing the encoder to capture task-relevant state changes and action-conditioned dynamics. Second, instead of measuring similarity using pixel-level differences, we compute distances based on optical flow. This provides a motion-aware measure of state similarity that is less sensitive to pixel-level noise and appearance variations. As a result, the contrastive objective emphasizes meaningful changes in scene dynamics while discounting irrelevant visual details. Combined with latent prediction and SIGReg, our motion-guided contrastive regularization encourages representations that are both statistically non-degenerate and sensitive to the dynamics required for control. The resulting world model learns task-relevant latent representations without relying on pixel reconstruction losses or auxiliary action-decoding objectives.

We evaluate our framework on the OGBench~\citep{ogbench} and LIBERO~\citep{libero} benchmarks under varying levels of visual complexity. Through rigorous representation analysis and downstream planning evaluations, our experiments reveal that while static collapse fundamentally impairs standard SIGReg-regularized models, our proposed contrastive objective effectively circumvents this failure. Ultimately, our approach yields substantial improvements in planning performance amid increased background diversity and scene clutter.
As a result, our method substantially boosts planning robustness under challenging visual conditions, increasing success rates from 52\% to 78\% on OGBench-Cube with randomized backgrounds and from 14\% to 54\% on LIBERO-Object compared to LeWM.

This paper contributes to JEPA world models as follows. First, we propose a motion-guided contrastive objective that explicitly preserves action-conditioned state changes in the learned latent space. Second, we introduce an optical-flow-based similarity measure that focuses the contrastive objective on meaningful scene dynamics rather than low-level visual differences. Third, we show across multiple benchmark datasets that our approach mitigates static-background-induced representation collapse and substantially improves planning performance under visually complex conditions.

\section{Related Work}
\label{sec:related_work}
\noindent \textbf{Latent world models.} Latent world models learn compact environment dynamics to support planning and policy learning. World Models~\citep{worldmodels}, PlaNet~\citep{planet}, Dreamer and DreamerV3~\citep{dreamer,dreamerv3}, and IRIS~\citep{transformerWM} learn to model dynamics and control in latent space, while supervising the process with observation reconstruction. MuZero~\citep{MuZero}, EfficientZero and EfficientZero V2~\citep{efficientzero,efficientzerov2}, TD-MPC~\citep{TD-MPC}, and TD-MPC2~\citep{TD-MPC2} learn reconstruction-free, task-oriented models whose representations are shaped by reward and value supervision. DreamerPro~\citep{dreamerpro}, MuDreamer~\citep{mudreamer}, and Dreamer-CDP~\citep{dreamercdp} also combine reconstruction-free representation learning with reward-driven control.

JEPA methods, including I-JEPA~\citep{ijepa} and V-JEPA~\citep{vjepa}, learn visual representations by predicting target embeddings without reconstructing pixels. For control, DINO-WM~\citep{dinowm}, V-JEPA 2-AC~\citep{vjepa2}, and JEPA-WM~\citep{jepawms} learn action-conditioned dynamics over pretrained visual representations, enabling image-goal planning without task-reward supervision. PLDM~\citep{pdlm} and LeWM~\citep{lewm} jointly learn the encoder and dynamics predictor from offline trajectories without observation reconstruction or task rewards. We focus on this end-to-end setting, where representations and dynamics can be learned together from offline interactions for image-goal planning. However, under joint optimization, latent prediction alone admits collapsed solutions, making the preservation of informative representations a central challenge. 

\noindent \textbf{Representation collapse and anti-collapse regularization.}
Anti-collapse regularization promotes representation diversity during joint encoder--predictor training. VICReg~\citep{vicreg} constrains feature variance and covariance, while SIGReg~\citep{lejepa}, adopted by LeWM~\citep{lewm}, encourages an isotropic Gaussian embedding distribution. However, statistical diversity alone does not ensure that representations preserve control-relevant information. Prior work has shown that JEPA models can favor static background features while failing to capture moving-object states~\citep{slowfeatures}. Other approaches address this issue through task- or action-based supervision. Bisimulation-based encoders~\citep{bisimularInvariant} encourage representations that preserve behaviorally relevant distinctions while remaining invariant to distractors. PLDM~\citep{pdlm} uses inverse-dynamics supervision to retain action information. Concurrent with our work, Sensorimotor World Models~\citep{sensormoterworld} and Delta-JEPA~\citep{delta-jepa} also explore inverse-dynamics supervision for learning action-sensitive representations. Our method instead introduces a within-episode contrastive objective guided by motion similarity. It encourages predicted latent transitions to distinguish alternative outcomes while preserving similarities between related transitions, without requiring an auxiliary action decoder.

% Another approach encourages representations to retain information about executed actions. PLDM~\citep{pdlm} incorporate inverse-dynamics objectives with VICReg loss. We also noticed concurrent works on combining inverse-dynamics objectives with SIGReg loss such as SMWM~\citep{sensormoterworld} and Delta-JEPA~\citep{delta-jepa}. \todo{Yiwen: is this transition too abrupt?} Our work examines static collapse in SIGReg-regularized manipulation world models, where diversity across episodes can coexist with weak sensitivity to manipulation state within an episode. We complement distributional regularization with motion-guided contrastive supervision, using similarities between observed motions to structure latent transitions.

\noindent \textbf{Contrastive and motion-guided representation learning}. Contrastive objectives enable latent prediction without requiring observation reconstruction~\citep{cpc,cswm}. Building on this idea, recent approaches apply contrastive supervision to action-conditioned prediction~\citep{taco,twister} and the learning of physical dynamics~\citep{dysib}. The choice of negative samples also influences the learned representation: \citet{contrastInductiveBias} show that within-trajectory negatives suppress shortcuts based on trajectory-specific static noise and improve the recovery of underlying dynamical variables. In video representation learning, synthetic motion augmentations and complementary visual-motion objectives have been used to encourage motion-sensitive features~\citep{Tubelet-Contrastive,CoCLR,MaCLR}. Most closely related to our approach, \citet{opticalFlowsimilarity} supervise representations by aligning pixel-embedding similarity distributions with targets derived from optical-flow similarity. Our method combines within-episode contrastive sampling with motion-derived soft targets to explicitly structure action-conditioned latent transitions.

% Contrastive objectives support latent prediction without observation reconstruction~\citep{cpc,cswm}. Recent methods apply contrastive supervision to action-conditioned prediction~\citep{taco,twister} and physical dynamics~\citep{dysib}. Negative selection also shapes the information retained: \citet{contrastInductiveBias} show that within-trajectory negatives can suppress shortcuts based on trajectory-specific static noise and improve recovery of dynamical variables. Video representation methods use synthetic motion and complementary visual and motion features to encourage motion-sensitive representations~\citep{Tubelet-Contrastive,CoCLR,MaCLR}. More directly related to our supervision, \citet{opticalFlowsimilarity} align pixel-embedding similarity distributions with targets derived from optical-flow similarities. Our method combines within-episode sampling with motion-derived soft targets to structure action-conditioned latent transitions.

% Contrastive learning supports latent prediction without reconstructing observations. CPC~\citep{CPC} learns representations by contrasting future samples against negatives, while C-SWM~\citep{cswm} uses contrastive training to learn structured latent dynamics. Negative selection also shapes the information retained: ~\citet{contrastInductiveBias} show that sampling negatives within a trajectory can suppress shortcuts based on trajectory-specific static noise and improve recovery of dynamical variables.

\section{Preliminaries and static collapse}
\label{sec:preliminaries}

\noindent \textbf{LeWorldModel.} We build on LeWM~\citep{lewm}, a JEPA-style world model trained end-to-end from visual observations. Given an observation $\mathbf{x}_t$ and action $\mathbf{a}_t$, its ViT encoder~\citep{vit} and transformer predictor compute
\begin{equation}
\mathbf{z}_t = \mathrm{enc}_{\theta}(\mathbf{x}_t), \qquad
\hat{\mathbf{z}}_{t+1} = \mathrm{pred}_{\phi}(\mathbf{z}_t, \mathbf{a}_t).
\label{eq:lewm}
\end{equation}
For clarity, we write the predictor in single-step form; in practice, it conditions on the embeddings of the three most recent observations and their associated actions (Appendix~\ref{app:experiments}).

The prediction is supervised by the encoded next observation
$\mathbf{z}_{t+1} = \mathrm{enc}_{\theta}(\mathbf{x}_{t+1})$:
\begin{equation}
\mathcal{L}_{\mathrm{pred}}
= \mathbb{E}\!\left[
\lVert \hat{\mathbf{z}}_{t+1} - \mathbf{z}_{t+1} \rVert_2^2
\right],
\label{eq:pred}
\end{equation} 
where the expectation is over training transitions. To discourage constant collapse, LeWM additionally uses SIGReg~\citep{lejepa}, which encourages minibatch embeddings $\mathbf{Z}$ to follow an isotropic Gaussian distribution. The encoder and predictor are jointly optimized with
\begin{equation}
\mathcal{L}_{\mathrm{LeWM}}
= \mathcal{L}_{\mathrm{pred}}
+ \lambda\,\mathrm{SIGReg}(\mathbf{Z}),
\label{eq:total}
\end{equation} where $\lambda$ controls the regularization strength. Although this objective encourages prediction accuracy and batch-level diversity, it does not ensure that the representations preserve action-relevant state changes.

\noindent \textbf{Static collapse.} Although SIGReg is an effective regularizer and LeWM achieves promising results, it does not specify which information should be preserved in the latent representation. We identify a failure mode arising from this limitation: in environments with visually complex backgrounds, the encoder can satisfy the SIGReg objective by primarily encoding static scene appearance. Because background features already exhibit substantial diversity across observations, they provide an easy way to satisfy the Gaussianity constraint enforced by SIGReg, reducing the pressure to encode manipulation-relevant dynamics. Specifically, when background features alone are sufficient to achieve a low SIGReg loss, the learned representation tends to become dominated by those features:

 \begin{equation}
     \mathrm{SIGReg}(\mathbf{Z}_B) \approx 0 \;\Longrightarrow\; \mathbf{z}_t \approx \mathbf{Z}_B + \epsilon_t, \qquad \|\epsilon_t\| \ll \|\mathbf{Z}_B\|,
 \end{equation}

where $\mathbf{Z}_B$ denotes background features and $\epsilon_t$ captures the remaining dynamic information. Consequently, changes induced by robot-object interactions contribute only weakly to the representation, leading to poor discrimination between manipulation states and degraded downstream control performance.

We use the single-cube manipulation environment from OGBench~\citep{ogbench}, denoted \emph{Cube}, and construct a paired variant, \emph{Cube-RB}. For each episode in \emph{Cube-RB}, we sample a background image from ImageNet~\citep{imagenet} and keep it fixed throughout the episode. Only distant pixels ($depth \ge  1.0$\,m) are replaced, leaving the robot arm, cube, actions, and underlying dynamics unchanged. As a result, \emph{Cube} and \emph{Cube-RB} differ only in the presence of episode-specific background variation. We train LeWM on both datasets using identical hyperparameters, thereby isolating the effect of complex static visual content. Figure~\ref{fig:teaser}(B) visualizes the learned latent representations using t-SNE~\citep{van2008visualizing}. On \emph{Cube}, the embeddings distinguish different manipulation states within an episode. In contrast, on \emph{Cube-RB}, embeddings cluster primarily according to background identity, causing visually distinct manipulation states to collapse into compact episode-specific groups despite substantial changes in the robot and cube configurations. This representation collapse is accompanied by a significant drop in planning performance. As shown in Figure~\ref{fig:teaser}(C), LeWM's success rate decreases from 74\% on \emph{Cube} to 52\% on \emph{Cube-RB}.

This failure mode motivates our motion-guided contrastive objective. By contrasting candidate states within the same scene, the objective prevents the model from relying on static background cues and instead encourages representations that capture task-relevant dynamics.

\section{Motion-Guided Contrastive Regularization}
\label{sec:method}
\label{sec:contrastive_regularization}

We introduce our motion-guided contrastive objective in this section. We train an action-conditioned world model to distinguish the observed next state from alternative states within the same episode, using optical-flow similarity to assign soft contrastive targets. Comparing states within an episode reduces the usefulness of static background identity, while motion-based targets account for alternatives that represent similar transitions.

\begin{figure}[t]
 \centering
 \includegraphics[width=\linewidth]{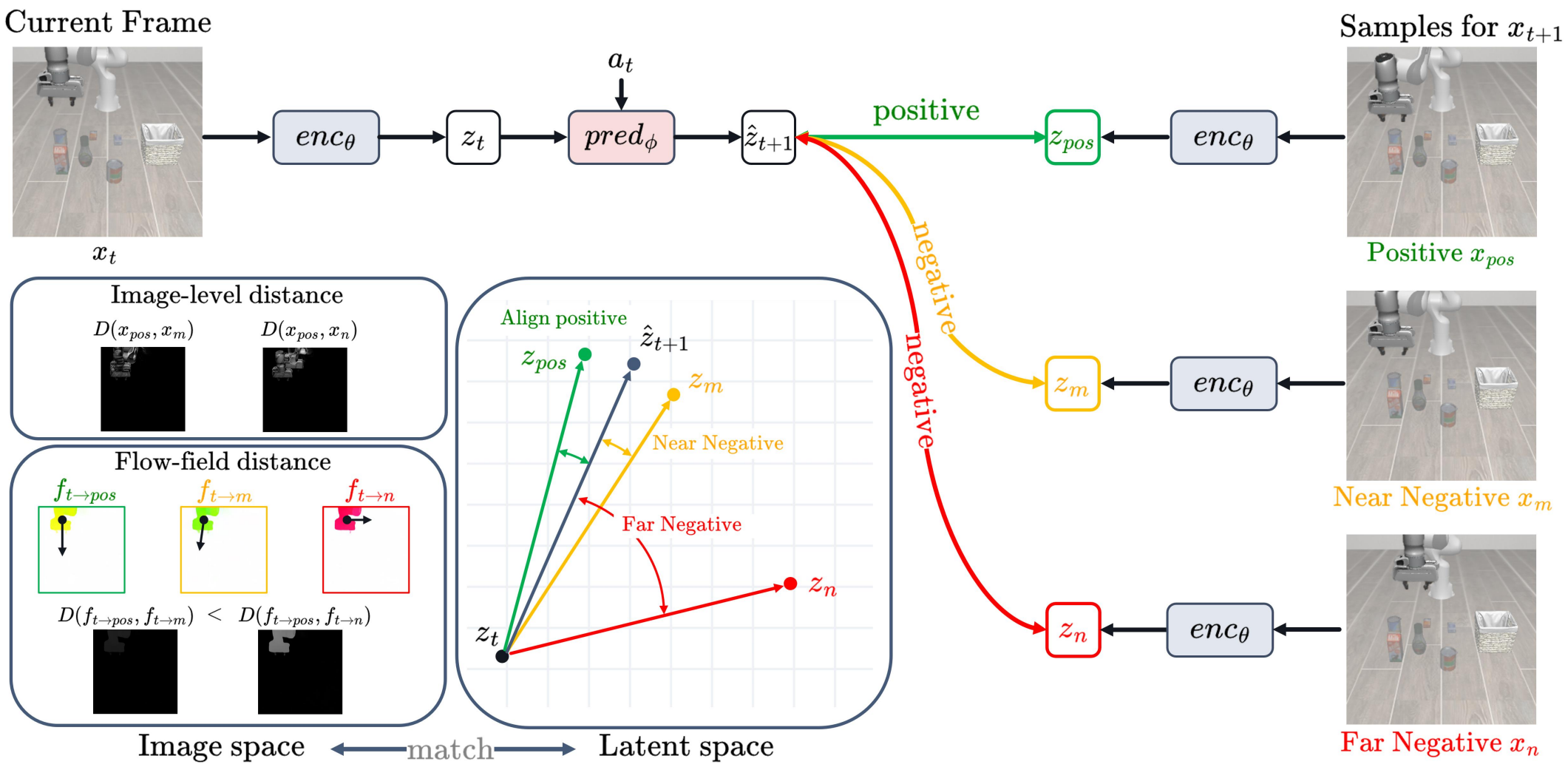}
 \caption{\textbf{Motion-guided contrastive learning.}
 The world model encodes the current observation $\mathbf{x}_t$ and predicts
 $\hat{\mathbf{z}}_{t+1}$.
 The true next observation $\mathbf{x}_{\mathrm{pos}}=\mathbf{x}_{t+1}$ provides the
 positive sample, while other frames from the same episode provide
 negative candidates $\mathbf{x}_m$ and $\mathbf{x}_n$.
 All observations share the same encoder.
 \textbf{Image-space:}
 Optical flow is computed from $\mathbf{x}_t$ to each candidate.
 Comparing these fields with the positive flow assigns higher target
 similarity to candidates with similar motion ($\mathbf{x}_m$) than to those
 with different motion ($\mathbf{x}_n$).
 \textbf{Latent-space:}
 Angles illustrate cosine similarities between the predicted
 transition $\hat{\mathbf{z}}_{t+1}-\mathbf{z}_t$ and candidate transitions $\mathbf{z}_j-\mathbf{z}_t$.
 Our motion-guided contrastive loss matches the resulting latent
 similarity distribution to the motion-derived target distribution
 through cross-entropy.
 The prediction loss aligns $\hat{\mathbf{z}}_{t+1}$ with
 $\mathbf{z}_{\mathrm{pos}}$. The grippers in the sample frames are highlighted for illustration purposes only. 
 }
 \label{fig:method}
\end{figure}

\textbf{Within-episode candidates.} Given an observed transition $(\mathbf{x}_t,\mathbf{a}_t,\mathbf{x}_{t+1})$, we use the next observation $\mathbf{x}_{t+1}$ as the positive sample and sample $N$ additional observations from the same episode. Let $\mathcal{N}_t$ contain their time indices, excluding $t$ and $t+1$, and let $\mathcal{C}_t=\{t+1\}\cup\mathcal{N}_t$ index all candidates. These observations share the same scene but may differ in the configuration of the robot and manipulated objects. Static background identity alone therefore cannot distinguish the positive from the other candidates.

\textbf{Latent transition similarities.} A shared encoder maps each observation to a latent state, $\mathbf{z}_i=\mathrm{enc}_{\theta}(\mathbf{x}_i)$, and an action-conditioned predictor estimates the next latent state, $\hat{\mathbf{z}}_{t+1}=\mathrm{pred}_{\phi}(\mathbf{z}_t,\mathbf{a}_t)$. We express the prediction and each candidate as changes from the current latent state:

\begin{equation}
    \widehat{\Delta\mathbf{z}}_t
    =\hat{\mathbf{z}}_{t+1}-\mathbf{z}_t,
    \qquad
    \Delta\mathbf{z}_{t,j}
    =\mathbf{z}_j-\mathbf{z}_t,
    \quad j\in\mathcal{C}_t.
\end{equation}

Inspired by \citet{opticalFlowsimilarity}, who match pixel embedding-similarity distributions to optical-flow-derived targets, we adopt a distribution-matching formulation for latent transitions. Using the predicted latent change $\widehat{\Delta z}_t$ as the contrastive anchor, we apply a softmax with temperature $\tau_z$ to its cosine similarities with the candidate changes $\Delta z_{t,j}$:

\begin{equation}
    p_{t,j}
    =
    \frac{
        \exp\!\left(
            \mathrm{sim}(
                \widehat{\Delta\mathbf{z}}_t,
                \Delta\mathbf{z}_{t,j}
            )/\tau_z
        \right)
    }{
        \sum_{k\in\mathcal{C}_t}
        \exp\!\left(
            \mathrm{sim}(
                \widehat{\Delta\mathbf{z}}_t,
                \Delta\mathbf{z}_{t,k}
            )/\tau_z
        \right)
    },
\end{equation}

where $\mathrm{sim}(\cdot, \cdot)$ denotes cosine similarity and $\tau_z>0$ is a temperature. This distribution describes how closely each candidate's latent change matches the predicted change.

\textbf{Motion-guided supervision.} Other observations in the episode may depict transitions similar to the observed next transition. As shown in Figure~\ref{fig:method}, image-level distance $D(\mathbf{x}_{pos}, \mathbf{x}_m)$ can be sensitive to irrelevant appearance variations. We therefore assign soft targets based on their motion similarity, rather than treating every alternative as equally dissimilar. 

For each candidate, we compute an optical-flow field from the current observation:
\begin{equation}
    \mathbf{f}_{t \rightarrow j}
    =\mathrm{Flow}(\mathbf{x}_t,\mathbf{x}_j),
    \qquad j\in\mathcal{C}_t.
\end{equation}
We compute optical flow with a pretrained SEA-RAFT model~\citep{wang2024sea}; settings are given in Appendix~\ref{app:implementation}.

To prevent large static regions from diluting motion differences, we weight each location $u$ by the larger flow magnitude in the positive and candidate transitions:
\begin{equation}
    w_{t,j}(u)
    =
    \max\!\left(
        \|\mathbf{f}_{t \rightarrow t+1}(u)\|_2,
        \|\mathbf{f}_{t \rightarrow j}(u)\|_2
    \right).
\end{equation}
We then measure motion dissimilarity $d_{t, j}$ for all pixels $\Omega$ using a weighted endpoint distance:
\begin{equation}
    \label{eq:flow_distance}
    d_{t,j}
    =
    \frac{
        \sum_{u\in\Omega}
        w_{t,j}(u)
        \|\mathbf{f}_{t \rightarrow t+1}(u)-\mathbf{f}_{t \rightarrow j}(u)\|_2
    }{
        \sum_{u\in\Omega}w_{t,j}(u)+\epsilon
    },
\end{equation}
where $\epsilon>0$ ensures numerical stability. Locations with little motion in both transitions receive little weight, while motion in either transition increases their contribution. The vector distance captures differences in both displacement magnitude and direction.

These distances define the target distribution:
\begin{equation}
    q_{t,j}
    =
    \frac{\exp(-d_{t,j}/\tau_f)}
    {\sum_{k\in\mathcal{C}_t}\exp(-d_{t,k}/\tau_f)},
\end{equation}
where $\tau_f>0$ controls target sharpness. The positive has zero distance and therefore receives maximal target probability, potentially tied with candidates that have identical flow. Other candidates receive probability according to their motion similarity to the positive.

The motion-guided contrastive loss matches the latent similarity distribution to these targets:
\begin{equation}
    \mathcal{L}_{\mathrm{MGC}}
    =
    -\mathbb{E}_{t,\mathcal{N}_t}
    \left[
        \sum_{j\in\mathcal{C}_t}
        q_{t,j}\log p_{t,j}
    \right],
\end{equation}
where the expectation is over observed transitions and sampled candidate sets. A one-hot target on $t+1$ would recover a single-positive InfoNCE loss with the same latent transition scores. Our soft targets instead encourage the predicted latent change to remain compatible with alternatives whose observed motion resembles the positive transition.

\textbf{Training objective.} We retain LeWM's latent prediction loss and SIGReg regularization and add the motion-guided contrastive loss:
\begin{equation}
    \mathcal{L}
    =
    \mathcal{L}_{\mathrm{pred}}
    +
    \lambda\,\mathrm{SIGReg}(\mathbf{Z})
    +
    \beta\,\mathcal{L}_{\mathrm{MGC}},
    \label{eq:overall_objective}
\end{equation}
where $\mathbf{Z}$ denotes the embeddings used for regularization in a training batch, and $\lambda,\beta\geq 0$ control the two regularization terms. The prediction loss learns action-conditioned dynamics, SIGReg regularizes the embedding distribution, and our contrastive loss encourages latent transitions to distinguish candidates according to their observed motion.

\section{Experiments}
\label{sec:experiments}

\begin{table}[t]
\caption{Planning success rate (\%) on six settings. $^{\dagger}$ uses a frozen pretrained visual encoder. Peak GPU memory and planning time are measured on Cube dataset  with an A5000 GPU and averaged over its 50 evaluation instances; planning time is the wall-clock time to plan one full trajectory (five replanning rounds). CEM uses 128 candidates and 10 iterations per round, with a planning horizon of up to 5 control steps.}    
\label{tab:main_results}
    \begin{center}
    \footnotesize
    \setlength{\tabcolsep}{4pt}
    % two-line header cell; method name with a smaller citation on the same line
    \newcommand{\hd}[1]{\begin{tabular}{@{}c@{}}#1\end{tabular}}
    \newcommand{\mth}[2]{#1~{\scriptsize\citep{#2}}}
    \resizebox{\linewidth}{!}{%
    \begin{tabular}{lcccc|cc|cc}
    \toprule
    \multicolumn{1}{c}{\hd{\textbf{Method}}} & \textbf{Push-T} & \textbf{2-Rooms} & \textbf{Reacher} & \textbf{Cube} & \textbf{Cube-RB} & \hd{\textbf{LIBERO-}\\\textbf{Object}} & \hd{\textbf{Peak GPU}\\\textbf{mem.\ (MB)}} & \hd{\textbf{Planning}\\\textbf{time (ms)}} \\
    \midrule
    \mth{DINO-WM$^{\dagger}$}{dinowm} & 74 & \textbf{100} & 78 & 78 & 68 & \textbf{68} & 20{,}714 & 11{,}223 \\
    \mth{PLDM}{pdlm} & 78 & 97 & 64 & 65 & 62 & 38 & 338 & 28 \\
    \mth{SMWM}{sensormoterworld} & 83 & 99 & 66 & \textbf{84} & \textbf{78} & 40 & 374 & 101 \\
    \mth{LeWM}{lewm} & \textbf{90} & 86 & \textbf{80} & 74 & 52 & 14 & 372 & 112 \\
    Ours & \textbf{90} & 88 & 78 & 76 & \textbf{78} & 54 & 372 & 112 \\
    % \hline
    % \rowcolor{myrowcolor} 
    % Ours & \textbf{90} & 88 & 78 & 76 & \textbf{78} & 54 & 372 & 112 \\
    % \midrule
    % Simulator rollouts & -- & 100 & -- & 94 & 94 & 80 & -- & -- \\
    \bottomrule
    \end{tabular}%
    }
    \end{center}
    \end{table}

\subsection{Experimental Setup}
\label{sec:experimental_setup}

\noindent\textbf{Benchmark datasets.}
We evaluate on six visual-control settings. The controlled pair introduced in Section~\ref{sec:preliminaries}, \emph{Cube}~\citep{ogbench} and \emph{Cube-RB} (Cube with randomized backgrounds), isolates the effect of episode-specific background variation. We additionally evaluate on LIBERO-Object~\citep{libero} to assess whether the benefits extend to multi-object scenes without synthetic background replacement. For LIBERO-Object, we expand the training data with noise-perturbed expert trajectories. Appendix~\ref{app:libero_data} details this data preparation and evaluation protocol. Finally, we evaluate on Push-T~\citep{dinowm}, 2-Rooms~\citep{pdlm}, and Reacher~\citep{reacher}, following prior world-model evaluations~\citep{dinowm,lewm}, to assess whether our objective maintains performance in visually simpler environments.

\noindent\textbf{Baseline models.}
We compare against four learned world-model baselines. \textbf{LeWM}~\citep{lewm} is our backbone, trained with its original prediction and SIGReg objectives without the proposed contrastive loss. \textbf{Sensorimotor World Models} (SMWM)~\citep{sensormoterworld} encourages action-sensitive representations through an auxiliary action-decoding objective. \textbf{PLDM}~\citep{pdlm} combines latent prediction with VICReg-inspired regularization and an auxiliary inverse-dynamics objective. \textbf{DINO-WM}~\citep{dinowm} learns dynamics over frozen DINOv2 features~\citep{dinov2}, providing a reference based on large-scale visual pretraining; unlike the other methods, it is not trained end-to-end on task data, and we mark it with $^{\dagger}$ in Table~\ref{tab:main_results}.

% We additionally report \textbf{Simulator rollouts}, a reference condition that replaces the learned world model with ground-truth simulator rollouts and thus plans with accurate dynamics.

\noindent\textbf{Training details.}
All methods are trained on the same task data and observation augmentations for 100 epochs, using a batch size of 128 and a learning rate of $5\times10^{-5}$. LeWM and our method share identical architectures and optimization settings, differing only in the addition of our motion-guided contrastive objective. Following LeWM~\citep{lewm}, we use $M=1024$ random projections for SIGReg and set its regularization weight to $\lambda=0.09$. We compute optical flow using SEA-RAFT~\citep{wang2024sea}. For each training transition, we sample $N=6$ additional candidate observations from the same episode. We set the flow temperature to $\tau_f=20$, the latent temperature to $\tau_z=0.1$, and the contrastive loss weight to $\beta=0.3$. We analyze sensitivity to the temperatures $\tau_f$ and $\tau_z$, the number of additional candidates $N$, the SIGReg weight $\lambda$, and the contrastive loss weight $\beta$ in Appendix~\ref{app:additional_results}.

\noindent\textbf{Evaluation via latent planning.}
We evaluate the learned world models through image-goal planning using a shared, sampling-based model-predictive control (MPC) procedure. Following LeWM~\citep{lewm}, we encode the current and goal observations and predict latent rollouts for candidate action sequences. The Cross-Entropy Method (CEM)~\citep{rubinstein1999cross} optimizes these sequences by minimizing the squared Euclidean distance between the predicted terminal embedding and the goal embedding. Holding the controller fixed across methods evaluates how well their learned representations and dynamics support goal-directed control. Our comparison focuses on offline, reward-free world models used within this common planning framework. To compare computational cost, we also report the peak GPU memory (\emph{Peak GPU mem.}) and the wall-clock planning time (\emph{Planning time}) required to plan one trajectory. Appendix~\ref{app:mpc_eval} provides the planning configuration and evaluation protocol.

\begin{figure}[!t]
% \begin{figure}[t]
\centering
% Rows: original frame, LeWM reconstruction, our reconstruction (labelled once
% on the left); three dataset blocks side by side, one column per sample.
% All images share the same square size \decw.
\newlength{\decsep}\setlength{\decsep}{0.004\linewidth}% gap between images
\newlength{\decgap}\setlength{\decgap}{0.008\linewidth}% gap between datasets
% Horizontal row labels: fixed line breaks (never re-wrap), and the label column
% is sized from the widest label so it can never overlap the images.
\newcommand{\declabtext}[1]{{\scriptsize\begin{tabular}{@{}c@{}}#1\end{tabular}}}
\newsavebox{\declabbox}\sbox{\declabbox}{\declabtext{Source\\images}}
\newlength{\declabw}\setlength{\declabw}{\wd\declabbox}
\newlength{\decw}\setlength{\decw}{\dimexpr(\linewidth-\declabw-2\decsep-6\decsep-2\decgap-4pt)/9\relax}
% label box exactly as tall as an image (no depth), text centered in it
\newcommand{\declabel}[1]{\leavevmode\vbox to\decw{\vfil\hbox to\declabw{\hfil\declabtext{#1}\hfil}\vfil}}
% rows stacked with exactly \decsep between them, matching the column gap
\newcommand{\decstack}{\baselineskip=0pt\lineskiplimit=0pt\lineskip=\decsep}
\begin{minipage}[t]{\declabw}
\centering\decstack
{\small\strut}\\
\declabel{Source\\images}\\
\declabel{LeWM}\\
\declabel{Ours}
\end{minipage}\hspace{2\decsep}%
\begin{minipage}[t]{\dimexpr3\decw+2\decsep+1pt\relax}
\centering\decstack
{\small\textbf{Cube}\strut}\\
\includegraphics[width=\decw,height=\decw]{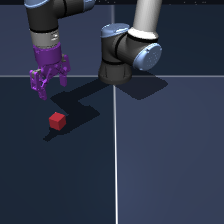}\hspace{\decsep}\includegraphics[width=\decw,height=\decw]{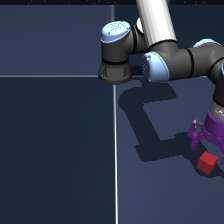}\hspace{\decsep}\includegraphics[width=\decw,height=\decw]{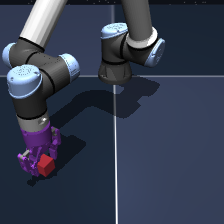}\\
\includegraphics[width=\decw,height=\decw]{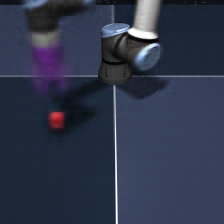}\hspace{\decsep}\includegraphics[width=\decw,height=\decw]{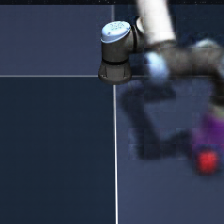}\hspace{\decsep}\includegraphics[width=\decw,height=\decw]{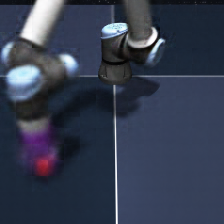}\\
\includegraphics[width=\decw,height=\decw]{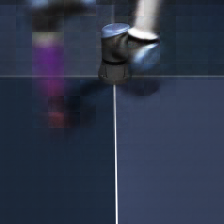}\hspace{\decsep}\includegraphics[width=\decw,height=\decw]{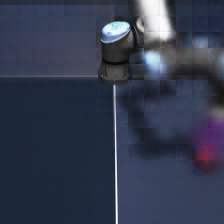}\hspace{\decsep}\includegraphics[width=\decw,height=\decw]{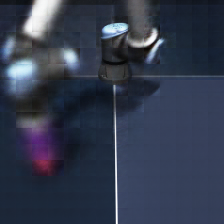}
\end{minipage}\hspace{\decgap}%
\begin{minipage}[t]{\dimexpr3\decw+2\decsep+1pt\relax}
\centering\decstack
{\small\textbf{Cube-RB}\strut}\\
\includegraphics[width=\decw,height=\decw]{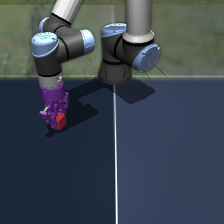}\hspace{\decsep}\includegraphics[width=\decw,height=\decw]{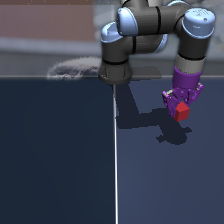}\hspace{\decsep}\includegraphics[width=\decw,height=\decw]{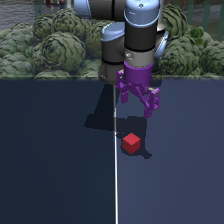}\\
\includegraphics[width=\decw,height=\decw]{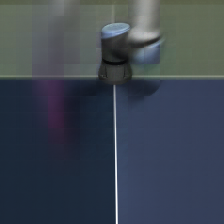}\hspace{\decsep}\includegraphics[width=\decw,height=\decw]{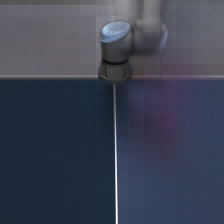}\hspace{\decsep}\includegraphics[width=\decw,height=\decw]{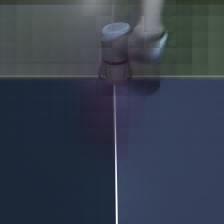}\\
\includegraphics[width=\decw,height=\decw]{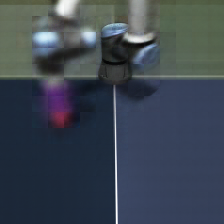}\hspace{\decsep}\includegraphics[width=\decw,height=\decw]{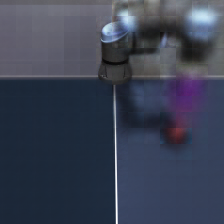}\hspace{\decsep}\includegraphics[width=\decw,height=\decw]{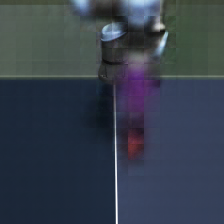}
\end{minipage}\hspace{\decgap}%
\begin{minipage}[t]{\dimexpr3\decw+2\decsep+1pt\relax}
\centering\decstack
{\small\textbf{LIBERO-Object}\strut}\\
\includegraphics[width=\decw,height=\decw]{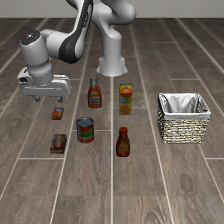}\hspace{\decsep}\includegraphics[width=\decw,height=\decw]{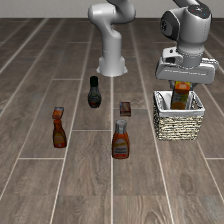}\hspace{\decsep}\includegraphics[width=\decw,height=\decw]{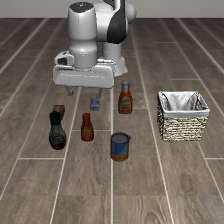}\\
\includegraphics[width=\decw,height=\decw]{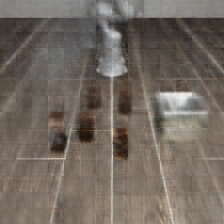}\hspace{\decsep}\includegraphics[width=\decw,height=\decw]{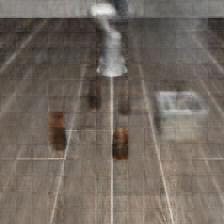}\hspace{\decsep}\includegraphics[width=\decw,height=\decw]{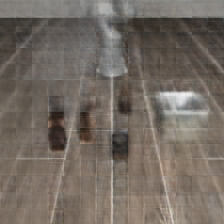}\\
\includegraphics[width=\decw,height=\decw]{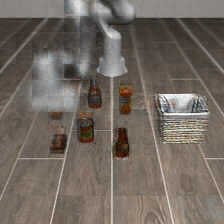}\hspace{\decsep}\includegraphics[width=\decw,height=\decw]{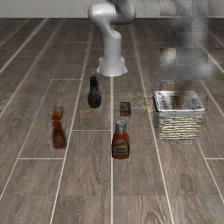}\hspace{\decsep}\includegraphics[width=\decw,height=\decw]{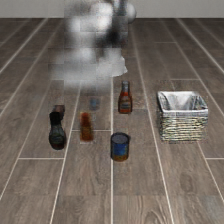}
\end{minipage}
\vspace{-0.3em}
\caption{Latent-space decoding on Cube, Cube-RB, and LIBERO-Object. Compared to LeWM, reconstructions from our frozen latents more clearly preserve the robot and manipulated objects on the challenging Cube-RB and LIBERO-Object datasets. Check our video demo for more comparisons. 
% \qiqi{move this figure to the top of the page}
}
\label{fig:decoder}
\vspace{-1em}
\end{figure}

% Each column is a different sample; rows show the source image and reconstructions from LeWM and our latents.

\subsection{Main Results}
\label{sec:main_results}

% 3x3 grid wrapped by text on the right: columns are datasets, rows are the
% input frame and the attention maps of LeWM and ours (labelled once on the left,
% as in Figure~\ref{fig:decoder}). All images share the same square size \attnw.
% Sizes are relative to \textwidth, since \linewidth changes inside wrapfigure.
\newlength{\attnsep}\setlength{\attnsep}{0.003\textwidth}% gap between images
\newlength{\attnw}\setlength{\attnw}{0.12\textwidth}% image size
% Horizontal row labels: fixed line breaks, column sized from the widest label.
\newcommand{\attnlabtext}[1]{{\scriptsize\begin{tabular}{@{}c@{}}#1\end{tabular}}}
\newsavebox{\attnlabbox}\sbox{\attnlabbox}{\attnlabtext{Source\\images}}
\newlength{\attnlabw}\setlength{\attnlabw}{\wd\attnlabbox}
\newlength{\attnfigw}\setlength{\attnfigw}{\dimexpr\attnlabw+2\attnsep+3\attnw+2\attnsep+1pt\relax}
% label box exactly as tall as an image (no depth), text centered in it
\newcommand{\attnlabel}[1]{\leavevmode\vbox to\attnw{\vfil\hbox to\attnlabw{\hfil\attnlabtext{#1}\hfil}\vfil}}
% Overlays are 224x225 px vs. 224x224 RGB frames: clip every image to the same
% 224x224 region (1 px = 1 bp for these PNGs) before scaling, so none is stretched.
\newcommand{\attntitle}[1]{\makebox[\attnw][c]{\scriptsize\textbf{#1}\strut}}
\begin{wrapfigure}{r}{\attnfigw}
\vspace{-1.2em}
\centering
% rows stacked with exactly \attnsep between them, matching the column gap
\begin{minipage}{\attnfigw}
\centering
\baselineskip=0pt\lineskiplimit=0pt\lineskip=\attnsep
\makebox[\attnlabw]{}\hspace{2\attnsep}\attntitle{Cube}\hspace{\attnsep}\attntitle{Cube-RB}\hspace{\attnsep}\attntitle{LIBERO-Object}\\
\attnlabel{Source\\images}\hspace{2\attnsep}\includegraphics[viewport=0 0 224 224,clip,width=\attnw,height=\attnw]{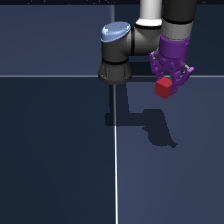}\hspace{\attnsep}\includegraphics[viewport=0 0 224 224,clip,width=\attnw,height=\attnw]{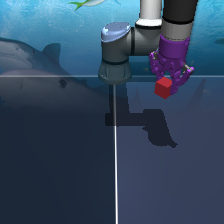}\hspace{\attnsep}\includegraphics[viewport=0 0 224 224,clip,width=\attnw,height=\attnw]{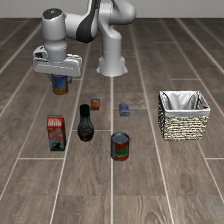}\\
\attnlabel{LeWM}\hspace{2\attnsep}\includegraphics[viewport=0 0 224 224,clip,width=\attnw,height=\attnw]{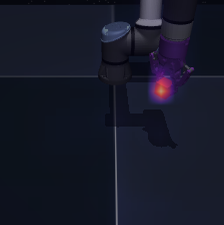}\hspace{\attnsep}\includegraphics[viewport=0 0 224 224,clip,width=\attnw,height=\attnw]{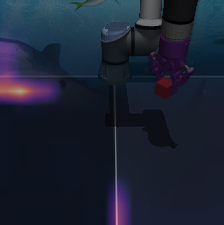}\hspace{\attnsep}\includegraphics[viewport=0 0 224 224,clip,width=\attnw,height=\attnw]{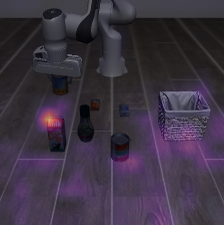}\\
\attnlabel{Ours}\hspace{2\attnsep}\includegraphics[viewport=0 0 224 224,clip,width=\attnw,height=\attnw]{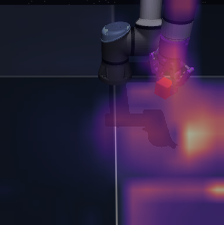}\hspace{\attnsep}\includegraphics[viewport=0 0 224 224,clip,width=\attnw,height=\attnw]{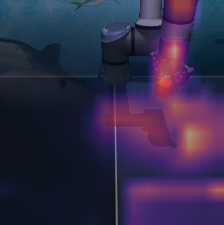}\hspace{\attnsep}\includegraphics[viewport=0 0 224 224,clip,width=\attnw,height=\attnw]{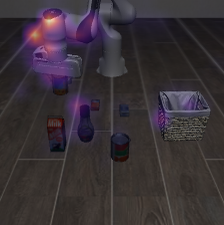}
\end{minipage}
\vspace{-0.6em}
\caption{Encoder attention over image patches (brighter = more attention). On Cube-RB and LIBERO-Object, LeWM attends to background and static objects; ours focuses on the gripper and manipulated object.}
\label{fig:attention_map}
\vspace{-1em}
\end{wrapfigure}

Table~\ref{tab:main_results} shows that motion-guided contrastive regularization substantially improves planning success over LeWM on Cube-RB and LIBERO-Object. On Cube-RB, success increases from 52\% to 78\%, a gain of 26 percentage points under background randomization. On LIBERO-Object, success under our image-goal planning protocol increases from 14\% to 54\%, extending the benefits to multi-object scenes, where the task-irrelevant static objects can cause similar static collapse behavior as in Cube-RB. On Cube, Push-T, 2-Rooms, and Reacher, our method's success rates remain on par. In particular, our method achieves 76\% on Cube and 78\% on Cube-RB, maintaining similar observed performance despite the introduction of episode-specific backgrounds, indicating the effectiveness of our method in preserving task-relevant information. Among the evaluated end-to-end world models, our method achieves the highest observed LIBERO-Object success rate. DINO-WM achieves higher success on this setting using its frozen pretrained visual encoder. On Cube-RB, our method matches SMWM at 78\%. These comparisons demonstrate substantial improvements over the LeWM backbone.

% \subsection{Analysis of the Learned Representation}
% \label{analysis}
%Include the decoder

\noindent\textbf{Decodability of manipulation state.}
% Following \citet{lewm}, we examine the information retained in the learned representations by training a lightweight transformer decoder to reconstruct observations from a single latent embedding. We train the decoder post hoc with the world-model parameters frozen. Its reconstructions therefore serve only as a diagnostic of the information retained by each representation.
Following \citet{lewm}, we train a lightweight transformer decoder to reconstruct observations from a single latent embedding. The world-model parameters remain frozen during decoder training, providing a qualitative probe of the visual information recoverable from each representation. Figure~\ref{fig:decoder} compares reconstructions on Cube, Cube-RB, and LIBERO-Object (ground truth, LeWM, ours). Reconstructions from LeWM latents retain the coarse scene layout but blur the manipulated object and end effector. Reconstructions from our latents more clearly depict object locations and gripper configurations. These results suggest that motion-guided contrastive supervision makes manipulation-relevant visual information more accessible to the decoder.

\noindent\textbf{Attention maps.}
We visualize the encoder's spatial focus when forming the global latent embedding by aggregating patch attention weights across all layers (Figure~\ref{fig:attention_map}). Rows show the input image and the attention maps for LeWM and our method. On Cube-RB and LIBERO-Object, LeWM assigns substantial attention to background regions and static objects, whereas our method focuses more strongly on the manipulated object and gripper. These patterns complement the reconstruction results and suggest greater emphasis on manipulation-relevant regions.

% We visualize the encoder's attention over image patches to examine its spatial focus when forming the global latent embedding (Figure~\ref{fig:attention_map}; columns show the input, LeWM, and ours attention maps, we visualize the heatmap of the cumulative attention weights over all layers). LeWM's attention is diffuse, with substantial weight on background regions and static objects. With motion-guided contrastive regularization, attention concentrates around the manipulated object and the gripper. These visualizations complement the decoder reconstruction results, suggesting increased attention to manipulation-relevant regions. However, attention weights alone do not establish which state information is retained in the latent representation.

\begin{wraptable}{r}{0.5\textwidth}
% \vspace{-1.2em}
\centering
\caption{MLP probing of physical state from frozen latents.
Position errors are Euclidean distances in centimeters.
Holding object (AUROC) is a binary gripper-state classification on LIBERO-Object.}
\label{tab:physical_probe}
\footnotesize
\setlength{\tabcolsep}{3pt}
\begin{tabular}{@{}llcc@{}}
\toprule
Dataset & Metric & LeWM & Ours \\
\midrule
\multirow{2}{*}{Cube}
 & Robot pos.\ error (cm) $\downarrow$ & 2.74 & \textbf{0.61} \\
 & Cube pos.\ error (cm) $\downarrow$ & 1.27 & \textbf{0.62} \\
\midrule
\multirow{2}{*}{Cube-RB}
 & Robot pos.\ error (cm) $\downarrow$ & 13.01 & \textbf{3.56} \\
 & Cube pos.\ error (cm) $\downarrow$ & 10.49 & \textbf{5.10} \\
\midrule
\multirow{2}{*}{\begin{tabular}{@{}l@{}}LIBERO-\\Object\end{tabular}}
 & Robot pos.\ error (cm) $\downarrow$ & 10.67 & \textbf{6.00} \\
 & Holding object (AUROC) $\uparrow$ & 0.78 & \textbf{0.88} \\
\bottomrule
\end{tabular}
\vspace{-1em}
\end{wraptable}

\noindent\textbf{Physical property probing.}
Qualitative reconstructions suggest that our latents retain manipulation-relevant visual information. To quantify physical-state recoverability, we train two-layer MLP probes on frozen encoder embeddings following~\citet{lewm}. We predict end-effector and cube positions on Cube and Cube-RB, and end-effector position and a binary object-holding indicator on LIBERO-Object. We report Euclidean position error in centimeters and the area under the receiver operating characteristic curve (AUROC) for object-holding classification. As shown in Table~\ref{tab:physical_probe}, our method improves physical-state recovery across all evaluated quantities. 

\FloatBarrier

% %Include the distance to goal like dinoWM
% \noindent\textbf{Alignment between latent cost and task progress}
% Latent-space planning requires more than accurate one-step prediction: the latent distance used by the planner must distinguish states that differ in their progress toward the goal. Static collapse can make this cost nearly constant within an episode, leaving CEM without a useful signal for ranking candidate action sequences.

% We test this prediction by recording the distance-to-goal cost \(C(\hat z_H)\) throughout action execution. Figure~\ref{fig:distance_to_goal} shows that, on clean OGBench-Cube, LeWM's cost decreases as the agent approaches the goal. On Cube-RB and LIBERO-Object, however, its cost changes little despite changes in the manipulation state, consistent with reduced sensitivity to task-relevant variation. With motion-guided contrastive regularization, the cost more clearly reflects task progress, decreasing as the goal is approached.
% \todo{Running out of space, maybe move this to appendix? It appears less straight-forward than the attention map and decoder reconstruction.}
% \input{figures/distance_to_goal}

\subsection{Ablation Studies}
\label{sec:ablation_studies}

\begin{wraptable}{r}{0.58\linewidth}
\vspace{-1.2em}
\centering
\small
\setlength{\abovecaptionskip}{4pt}
\setlength{\belowcaptionskip}{2pt}
\setlength{\tabcolsep}{4pt}
\renewcommand{\arraystretch}{1.15}
\caption{Ablation of training objectives on Cube-RB and LIBERO-Object.
Entries are planning success (\%). All variants retain the prediction loss.
InfoNCE uses the same within-episode candidates ($N = 6$) and latent transition scores ($\tau_z = 0.1$) as our method.}
\label{tab:ablation}
\begin{tabular*}{\linewidth}{@{\extracolsep{\fill}}clcc@{}}
\toprule
SIGReg & Contrastive objective & Cube-RB & LIBERO-Object \\
\midrule
$\checkmark$ & None          & 52 & 14 \\
$\checkmark$ & InfoNCE       & 56 & 34 \\
--           & Motion-guided & 58 & 14 \\
$\checkmark$ & Motion-guided & \textbf{78} & \textbf{54} \\
\bottomrule
\end{tabular*}
\end{wraptable}

We ablate contrastive supervision and SIGReg on Cube-RB and LIBERO-Object, retaining the prediction loss in all variants. As shown in Table~\ref{tab:ablation}, adding InfoNCE to LeWM increases planning success from 52\% to 56\% on Cube-RB and from 14\% to 34\% on LIBERO-Object. Our motion-guided objective further increases success to 78\% and 54\%, exceeding InfoNCE by 22 and 20 percentage points, respectively. These gains support the benefit of incorporating observed motion similarities into contrastive supervision. Removing SIGReg from the full objective reduces success from 78\% to 58\% on Cube-RB and from 54\% to 14\% on LIBERO-Object. These results indicate that motion-guided supervision and batch-level distributional regularization make complementary contributions. 

Additional ablations, including sensitivity to the temperatures $\tau_f$ and $\tau_z$, the number of additional candidates $N$, the SIGReg weight $\lambda$, and the contrastive loss weight $\beta$, are provided in Appendix~\ref{app:additional_results}.

% \qiqi{Remember to remove the appendix at the end of the paper. Also fix the appendix ref}

% \qiqi{Additional experimental details and ablations are provided in a separate appendix PDF (appendix.pdf), submitted as supplementary material alongside the main paper.}

% We ablate the training objectives on Cube-RB and LIBERO-Object to examine the contributions of contrastive supervision and SIGReg. As shown in Table~\ref{tab:ablation}, adding an InfoNCE objective to LeWM increases planning success from 14\% to 34\% on LIBERO-Object and from 52\% to 56\% on Cube-RB, indicating that contrastive supervision already improves upon prediction and distributional regularization alone. Our motion-guided objective further increases success to 54\% and 78\%, outperforming InfoNCE by 20 and 22 percentage points, respectively. This comparison supports the benefit of incorporating observed motion similarities into contrastive supervision.

% Removing SIGReg from our full objective reduces success to 14\% on LIBERO-Object and 58\% on Cube-RB. Motion-guided supervision and batch-level distributional regularization therefore play complementary roles: the proposed loss improves planning when combined with SIGReg but does not replace it. We examine sensitivity to $\tau_f$, $\tau_z$, $N$, and $\beta$ in Appendix~\ref{app:ablation}, varying each while holding the others fixed.
\section{Conclusion}
\label{sec:conclusion}
% Keep this section short (typically half a page or less).

We identified static collapse in JEPA-style world models, where representations preserve episode-specific static content while suppressing manipulation-relevant changes. We introduced motion-guided contrastive regularization, which uses optical flow to guide comparisons among observations from the same episode. Combined with latent prediction and SIGReg, it improves planning success from 52\% to 78\% on Cube-RB and from 14\% to 54\% on LIBERO-Object, with comparable performance on four other settings. Representation analyses and ablations support improved preservation of manipulation state and the complementary role of SIGReg.

\noindent \textbf{Limitation and future work}. 
Like LeWM, our method depends on sufficient action coverage and remains sensitive to shifts in action representations and visual appearance, including changes to LIBERO's rendering engine. It also assumes static backgrounds; background motion could distort the optical-flow similarities used for supervision. Future work should improve transfer across actions and appearances, isolate task-relevant motion in dynamic scenes, and extend planning to longer horizons.

% Required / recommended statements (do not count toward the page limit)
% \input{sections/reproducibility}

% Optional author contributions go here.

% Acknowledgments (including funding) go here for the camera-ready version.

\bibliography{references}
\bibliographystyle{iclr2027_conference}

\clearpage
\appendix
{\LARGE\sc Appendix\par}
\vspace{1.5em}
\section{Additional Experimental Details}
\label{app:experiments}

\subsection{LIBERO-Object Data Preparation}
\label{app:libero_data}

\noindent\textbf{Dataset partitioning.}
We use all tasks in LIBERO-Object, which provides 500 expert trajectories. Before generating additional data, we reserve three expert trajectories per task for evaluation and use the remaining trajectories for training. We generate perturbed continuations only from training trajectories, ensuring that held-out trajectories and their continuations are excluded from the training data.

\noindent\textbf{Perturbed trajectory generation.}
To increase training diversity around the demonstrated behaviors, we generate additional trajectories by perturbing expert actions. For each training trajectory, we uniformly sample three valid starting frames without replacement. At each starting frame, we restore the recorded simulator state and execute three perturbed continuations of up to 100 steps each, yielding nine rollouts per training demonstration.

We perturb the continuous pose components of the expert action sequence with zero-mean Gaussian noise and clip the resulting actions to $[-1,1]$. The three variants use noise standard deviations $\sigma \in \{0.15, 0.33, 0.5\}$, paired with per-step gripper-flip probabilities $p \in \{0.006, 0.03, 0.09\}$, respectively. For each continuation, the discrete gripper command is initialized from the expert command at the starting frame and subsequently undergoes stochastic sign flips with the corresponding probability. We record the executed actions and resulting simulator observations.

\noindent\textbf{Filtering and dataset composition.}
We discard a rollout if it contains fewer than five recorded steps, if the target object's vertical coordinate falls below $z=-0.60$ at any recorded step in simulator coordinates, or if the Euclidean displacement between its first and last recorded positions is less than $0.03\,\mathrm{m}$. We also tag distractor motion when the first-to-last change in selected static object-state coordinates has an $\ell_2$ norm greater than $0.05$. Rollouts exhibiting distractor motion are retained and assigned the tag \texttt{distractor\_moved=1}.

After filtering, we retain 4,108 perturbed trajectory chunks, which are combined with the training expert trajectories. All methods evaluated on LIBERO-Object use this same expanded training dataset, with one model per method trained jointly across all LIBERO-Object tasks.

\subsection{Latent Planning and Evaluation Protocol}
\label{app:mpc_eval}

\noindent\textbf{Planning objective.}
Given a current observation $\mathbf{x}_t$ and a goal observation $\mathbf{x}_g$, we compute their embeddings $\mathbf{z}_t=\mathrm{enc}_{\theta}(\mathbf{x}_t)$ and $\mathbf{z}_g=\mathrm{enc}_{\theta}(\mathbf{x}_g)$. For each candidate action sequence $\mathbf{a}=(\mathbf{a}_t,\ldots,\mathbf{a}_{t+H-1})$, the learned dynamics model autoregressively predicts a terminal embedding $\hat{\mathbf{z}}_{t+H}(\mathbf{a})$. We optimize the action sequence using the Cross-Entropy Method (CEM) to minimize the terminal goal-matching cost
\begin{equation}
    \mathcal{C}(\mathbf{a})
    =
    \left\|
    \hat{\mathbf{z}}_{t+H}(\mathbf{a})-\mathbf{z}_g
    \right\|_2^2.
\end{equation}
All methods use the same planning procedure and optimization budget, allowing us to compare how their learned representations and dynamics support goal-directed control.

\noindent\textbf{CEM and replanning.}
At each control step, CEM performs 10 optimization iterations, sampling 128 candidate action sequences per iteration. The 16 sequences with the lowest predicted costs form the elite set used to update the sampling distribution. One control step spans five environment timesteps, with observations sampled at this interval. Each evaluation instance has a total execution budget of five control steps, corresponding to 25 environment timesteps. After optimization, we execute only the first action of the selected sequence, obtain a new observation after one control step, and replan toward the same goal. The planning horizon matches the remaining execution budget, decreasing from five to one control step over successive replanning rounds: $H \in \{5,4,3,2,1\}$.

\noindent\textbf{Evaluation instances and checkpoints.}
For each benchmark setting, we evaluate 50 initial--goal pairs using evaluation seed 42. Each pair is constructed by sampling an initial timestep from a held-out trajectory and selecting the goal observation five control steps later in the same trajectory. For LIBERO-Object, the 50 instances span all ten tasks; this is the total evaluation count rather than the count per task. All methods are evaluated on the same initial--goal pairs, and we use the final saved training checkpoint for each method. Success rates are computed over the 50 evaluation instances in each benchmark setting.

\noindent\textbf{Success criteria.}
Success is determined by matching the underlying environment state to the sampled goal state using the following criteria:
\begin{itemize}
    \item \textbf{Push-T.}
    The combined Euclidean position error of the pusher and T-block must be less than 20 pixels in the native $512\times512$ coordinate system, and the wrapped T-block orientation error must be less than $\pi/9$ radians ($20^\circ$).

    \item \textbf{Reacher.}
    The absolute difference between each joint angle and its goal value must be less than $0.05$ radians, following the \texttt{qpos\_match} configuration.

    \item \textbf{2-Rooms.}
    The Euclidean distance between the agent's center and its goal position must be less than 16 pixels in the $224\times224$ environment coordinate system.

    \item \textbf{Cube and Cube-RB.}
    The Euclidean distance between the cube's 3D position and its goal position must be at most $0.04\,\mathrm{m}$. Cube orientation is excluded from the success criterion. Cube-RB uses the same criterion as Cube because background replacement changes only the observations, while preserving the underlying states and dynamics.

    \item \textbf{LIBERO-Object.}
    The maximum Euclidean position error across the task's movable objects of interest must be at most $0.05\,\mathrm{m}$ relative to their positions in the sampled goal state. Object orientations and robot joint errors are excluded. This criterion evaluates attainment of the sampled goal state rather than completion of the original LIBERO task objective.
\end{itemize}

\noindent\textbf{Initially satisfied goals.}
For Cube and Cube-RB, initial states that already satisfy the goal criterion are counted as successes, following the original LeWM~\citep{lewm} evaluation implementation. For LIBERO-Object, we reject initially satisfied initial--goal pairs and resample until 50 valid evaluation instances are obtained.

\subsection{Model and Training Implementation}
\label{app:implementation}

\noindent\textbf{Visual encoder.}
We use a ViT-Tiny/14 encoder trained from scratch, with 12 transformer blocks, 192-dimensional tokens, 3 attention heads, and 768-dimensional feed-forward layers with GELU. Each $224\times224$ RGB image is divided into 256 non-overlapping $14\times14$ patches, supplemented with a learned CLS token and learned positional embeddings. The final CLS feature passes through a projection head consisting of a linear layer from 192 to 2048 dimensions, BatchNorm, GELU, and a linear layer back to 192 dimensions.

\noindent\textbf{Temporal predictor.}
Observations are encoded independently and passed to an action-conditioned causal transformer with 6 blocks, 16 attention heads, 64 dimensions per head, and 2048-dimensional feed-forward layers. The predictor's input and output dimensions are both 192. Each training clip contains three context observations and one future observation, sampled at intervals of five environment timesteps. Using the subsampled time index of the main text, the predictor uses $(\mathbf{x}_{t-2},\mathbf{x}_{t-1},\mathbf{x}_t)$ and the associated actions to predict the embedding of $\mathbf{x}_{t+1}$.

\noindent\textbf{Optimization.}
LeWM and our method share the same architecture and optimization settings, differing only in the addition of the motion-guided contrastive objective. We train for 100 epochs with a batch size of 128 using AdamW, a learning rate of $5\times10^{-5}$, weight decay of $10^{-3}$, momentum coefficients $(0.9,0.999)$, and $\epsilon=10^{-8}$. The learning rate follows a linear warmup followed by cosine decay.

\noindent\textbf{Candidate sampling and motion supervision.}
We compute optical flow using the pretrained SEA-RAFT~\citep{wang2024sea} Spring-M checkpoint
\texttt{MemorySlices/Tartan-C-T-TSKH-spring540x960-M}.
The flow model is frozen and runs in evaluation mode with four refinement iterations and \texttt{scale=0}. Flow is computed at $224\times224$ resolution, with displacement vectors measured in pixels at that resolution. The offline pipeline converts stored RGB frames to channel-first floating-point tensors in $[0,255]$, which SEA-RAFT internally normalizes to $[-1,1]$. No additional resizing or augmentation is applied during flow precomputation for the native $224\times224$ frames.

For each anchor frame $\mathbf{x}_t$, we randomly sample 40 candidate frames from any other timestep of the same episode, excluding the current frame $\mathbf{x}_t$ and the positive frame $\mathbf{x}_{t+1}$, and precompute their motion-weighted distances $d_{t,j}$ (Eq.~\ref{eq:flow_distance}) from the anchor-to-positive and anchor-to-candidate flow fields. Only these scalar distances are cached, in float32 format; the dense flow fields are not retained. During training, we randomly draw $N$ of the 40 cached candidates for each transition ($N = 6$ by default) and construct the soft contrastive targets from their cached distances. SEA-RAFT inference is therefore performed entirely offline, and no flow model is run during world-model training.

\noindent\textbf{Training objective.}
We retain LeWM's latent prediction loss and SIGReg regularization, and add the motion-guided contrastive loss defined in the main text. SIGReg is applied to the pre-projection encoder embeddings of the current and next observations, using $M=1024$ random projections and regularization weight $\lambda=0.09$. Unless otherwise specified, the contrastive objective uses $\tau_f=20$, $\tau_z=0.1$, and $\beta=0.3$.

\subsection{Representation Analysis Implementation}
\label{app:representation_analysis}

\noindent\textbf{Frozen-representation evaluation.}
We assess the information retained by the learned representations through image reconstruction and physical-state probing. For each method, we freeze the world-model parameters and train auxiliary decoders or probes on the resulting embeddings. Only the auxiliary model is optimized during these analyses.

\noindent\textbf{Image reconstruction.}
We train a lightweight transformer decoder to reconstruct an observation from its corresponding latent embedding. These reconstructions provide a qualitative assessment of the visual information recoverable from each representation, particularly the configurations of the robot and manipulated objects.

On Cube and Cube-RB, both methods use the same episode-level partition, with 9,000 training episodes and 1,000 validation episodes, so all clips from a given episode remain within a single subset.

For LIBERO-Object, both methods use 47 source demonstrations per task for decoder training and three for validation. Perturbed continuations inherit the partition of their source demonstration, ensuring that each demonstration and its perturbed descendants remain in the same subset.

\noindent\textbf{Physical-state probing.}
We train two-layer MLP probes on frozen encoder embeddings to quantify the recoverability of manipulation-relevant state information. On Cube and Cube-RB, the probes predict the 3D positions of the end effector and cube. On LIBERO-Object, they predict the end-effector position and a binary object-holding indicator. We measure position-prediction errors using Euclidean distance in centimeters and evaluate object-holding classification using the area under the receiver operating characteristic curve (AUROC).

The numerical probing experiments use episode-level partitions into disjoint training, validation, and test subsets. For each of Cube and Cube-RB, these subsets contain 7,000, 1,500, and 1,500 episodes, respectively, corresponding to 56,000, 12,000, and 12,000 frames. For LIBERO-Object, they contain 350, 70, and 80 episodes, corresponding to 10,850, 2,160, and 2,487 frames. All frames from a given episode remain within a single subset.

\section{Additional Results}
\label{app:additional_results}

We examine the sensitivity of motion-guided contrastive regularization to the flow temperature $\tau_f$, latent temperature $\tau_z$, number of additional within-episode candidates $N$, SIGReg weight $\lambda$, and contrastive loss weight $\beta$. Each sweep varies one hyperparameter while keeping the remaining settings fixed, including the architecture, training schedule, and planning procedure. The reference settings are $\tau_f=20$, $\tau_z=0.1$, $N=6$, $\lambda=0.09$, and $\beta=0.3$. Table~\ref{tab:hyperparameter_sensitivity} summarizes planning success on LIBERO-Object under the evaluation protocol in Appendix~\ref{app:mpc_eval}.

\begin{table}[htbp]
    \centering
    \caption{
        Hyperparameter sensitivity on LIBERO-Object.
        ``Succ.'' denotes planning success (\%).
        Bold marks the highest result within each sweep;
        underlining marks the reference parameter value.
    }
    \label{tab:hyperparameter_sensitivity}

    \small
    \setlength{\tabcolsep}{3.5pt}
    \setlength{\arrayrulewidth}{0.3pt}
    \setlength{\aboverulesep}{0pt}
    \setlength{\belowrulesep}{0pt}
    \renewcommand{\arraystretch}{1.15}

    \begin{tabular}{@{}cc|cc|cc|cc|cc@{}}
        \toprule
        \multicolumn{2}{c|}{\textbf{(a) Flow temp.}} &
        \multicolumn{2}{c|}{\textbf{(b) Latent temp.}} &
        \multicolumn{2}{c|}{\textbf{(c) Candidates}} &
        \multicolumn{2}{c|}{\textbf{(d) SIGReg}} &
        \multicolumn{2}{c}{\textbf{(e) Contrastive}} \\

        \cmidrule(lr){1-2}
        \cmidrule(lr){3-4}
        \cmidrule(lr){5-6}
        \cmidrule(lr){7-8}
        \cmidrule(lr){9-10}

        $\tau_f$ & Succ. &
        $\tau_z$ & Succ. &
        $N$      & Succ. &
        $\lambda$ & Succ. &
        $\beta$  & Succ. \\

        \midrule
        2.5 & 40 &
        0.02 & 52 &
        2 & 52 &
        0 & 14 &
        0.05 & 50 \\

        5 & 36 &
        0.05 & \textbf{54} &
        4 & 48 &
        0.03 & 40 &
        0.1 & \textbf{54} \\

        10 & 46 &
        \underline{0.1} & \textbf{54} &
        \underline{6} & 54 &
        \underline{0.09} & \textbf{54} &
        \underline{0.3} & \textbf{54} \\

        \underline{20} & \textbf{54} &
        0.2 & 48 &
        8 & 50 &
        0.18 & \textbf{54} &
        0.6 & 52 \\

        40 & 40 &
        0.5 & 50 &
        10 & 54 &
        0.3 & 50 &
        1.0 & 38 \\

        80 & 34 &
        0.75 & 44 &
        12 & \textbf{58} &
        & &
        & \\

        \bottomrule
    \end{tabular}
\end{table}

\subsection{Flow Temperature $\tau_f$}
\label{app:ablation_tau_f}

The flow temperature $\tau_f$ controls the concentration of the motion-derived target distribution. Smaller values assign more probability to candidates with the lowest flow distances from the positive, whereas larger values distribute probability more evenly across candidates.

As shown in Table~\ref{tab:hyperparameter_sensitivity}(a), planning success reaches 54\% at $\tau_f=20$, the highest result among the tested settings. Success is lower at both smaller and larger temperatures, ranging from 36\% to 46\% for $\tau_f\in\{2.5,5,10\}$ and from 34\% to 40\% for $\tau_f\in\{40,80\}$. These results favor an intermediate temperature within the tested range. We use $\tau_f=20$ in the main experiments.

\subsection{Latent Temperature $\tau_z$}
\label{app:ablation_tau_z}

The latent temperature $\tau_z$ controls how cosine similarities between predicted and candidate latent transitions are mapped to a probability distribution. Smaller values produce sharper distributions, while larger values produce smoother distributions. Thus, $\tau_z$ controls the predicted distribution, whereas $\tau_f$ controls the motion-derived supervision targets.

Table~\ref{tab:hyperparameter_sensitivity}(b) shows that planning success ranges from 52\% to 54\% for $\tau_z\in\{0.02,0.05,0.1\}$, compared with 44\%--50\% for the larger tested values. The highest observed success is shared by $\tau_z=0.05$ and $\tau_z=0.1$. We use $\tau_z=0.1$ in the main experiments.

\subsection{Number of Additional Candidates $N$}
\label{app:ablation_N}

We vary the number of additional observations sampled from the same episode, evaluating $N\in\{2,4,6,8,10,12\}$ while fixing $\tau_f=20$ and $\tau_z=0.1$. Each comparison contains $N+1$ candidates, including the true next observation.

As shown in Table~\ref{tab:hyperparameter_sensitivity}(c), increasing $N$ does not yield a monotonic improvement in planning success. The highest observed success is 58\% at $N=12$, compared with 54\% at our default $N=6$. We retain $N=6$ to limit the number of additional observations encoded per training transition.

\subsection{SIGReg Weight $\lambda$}
\label{app:ablation_lambda}

We examine the strength of distributional regularization by varying the SIGReg weight $\lambda$ while fixing the contrastive loss weight at $\beta=0.3$. All configurations retain the prediction and motion-guided contrastive losses.

As shown in Table~\ref{tab:hyperparameter_sensitivity}(d), removing SIGReg yields 14\% planning success, consistent with the ablation in Table~\ref{tab:ablation}. Adding SIGReg increases success to 40\% at $\lambda=0.03$ and 54\% at both $\lambda=0.09$ and $\lambda=0.18$, while $\lambda=0.3$ yields 50\%. These results support the complementary role of distributional regularization alongside motion-guided supervision. We use $\lambda=0.09$ as the default.

\subsection{Contrastive Loss Weight $\beta$}
\label{app:ablation_beta}

We vary the weight $\beta$ of the motion-guided contrastive loss while fixing the SIGReg weight at $\lambda=0.09$. This experiment evaluates how the strength of motion-guided supervision affects planning performance.

Table~\ref{tab:hyperparameter_sensitivity}(e) shows that $\beta=0.1$ and $\beta=0.3$ both achieve the highest observed success of 54\%. The settings $\beta=0.05$ and $\beta=0.6$ yield 50\% and 52\%, respectively, whereas increasing the weight to $\beta=1.0$ reduces success to 38\%. Thus, assigning greater weight to the contrastive objective does not necessarily improve planning performance. We retain $\beta=0.3$ as the default.

\end{document}